\documentclass[a4paper,12pt]{article}

\usepackage{amsthm}
\usepackage{amssymb}
\usepackage{algorithm}
\usepackage{algorithmic}
\usepackage{balance}
\usepackage{url}
\usepackage{xspace}

\newcommand{\Reals}{\mathbb{R}}
\newcommand{\Nats}{\mathbb{N}}

\newcommand{\bigo}[1]{\mathord{O}\mathord{\left(#1\right)}}

\newcommand{\ie}{i.\,e.\xspace}
\newcommand{\eg}{e.\,g.\xspace}

\newcommand{\Space}{\mathcal{S}}

\newtheorem{theorem}{Theorem}
\newtheorem{lemma}{Lemma}
\newtheorem{proposition}{Proposition}

\begin{document}

\title{Evolutionary Algorithms and Dynamic Programming}


\author{Benjamin Doerr\thanks{Algorithms and Complexity, Max-Planck-Institut f{\"u}r Informatik, Saarbr{\"u}cken, Germany},
Anton Eremeev\thanks{Omsk Branch of Sobolev Institute of Mathematics SB RAS, Omsk, Russia},
Frank Neumann\thanks{School of Computer Science, University of Adelaide, Adelaide, Australia},\\
Madeleine Theile\thanks{Institut f{\"u}r Mathematik, TU Berlin, Berlin, Germany},
Christian Thyssen\thanks{Fakult{\"a}t f{\"u}r Informatik, LS 2, TU Dortmund, Dortmund, Germany} \footnote{n{\'e} Horoba}}

\maketitle

\begin{abstract}
Recently, it has been proven that evolutionary algorithms produce good results for a wide range of combinatorial optimization problems. Some of the considered problems are tackled by evolutionary algorithms that use a representation which enables them to construct solutions in a dynamic programming fashion. We take a general approach and relate the construction of such algorithms to the development of algorithms using dynamic programming techniques. Thereby, we give general guidelines on how to develop evolutionary algorithms that have the additional ability of carrying out dynamic programming steps. Finally, we show that for a wide class of the so-called DP-benevolent problems (which are known to admit FPTAS) there exists a fully polynomial-time randomized approximation scheme based on an evolutionary algorithm.
\end{abstract}






\section{Introduction}
\label{sec:introduction}

Evolutionary algorithms (EAs) \cite{EibSmi2007} have been shown to be successful for a wide range of optimization problems. While these algorithms work well for many optimization problems in practice, a satisfying and rigorous mathematical understanding of their performance is an important challenge in the area of evolutionary computing~\cite{BookAugDoe}.

Interesting results on the runtime behaviour of evolutionary algorithms have been obtained for a wide range of combinatorial optimization problems (see \cite{BookNeuWit} for a comprehensive presentation). This includes well-known problems such as sorting and shortest paths \cite{ScharnowSorting2004}, spanning trees \cite{journals/tcs/NeumannW07}, maximum matchings \cite{conf/stacs/GielW03}, and minimum cuts \cite{B10:nr-ammbemoa,B10:nrs-cmcbrsh}. There are also some results on evolutionary algorithms acting as approximation algorithms for NP-hard problems like partition \cite{conf/stacs/Witt05}, covering \cite{1277118}, and multi-objective shortest path \cite{Horoba2009,NeumannT10} problems. But a general theoretical explanation of the behavior of evolutionary algorithms is still missing. The first step in this direction is taken in \cite{B10:rs-eaamop-pr}, where the authors show for an important subclass of optimization problems that evolutionary algorithms permit optimal solutions in polynomial time.

\subsection{Main Contributions}

The aim of this paper is to make another contribution to the theoretical understanding of evolutionary algorithms for combinatorial optimization problems. We focus on the question how to represent possible solutions such that the search process becomes provably efficient. When designing an evolutionary algorithm for a given problem, a key question is how to choose a good representation of possible solutions. This problem has been extensively studied in the literature on evolutionary algorithms~\cite{BookRothlauf}; for example there are different representations for the well-known traveling salesman problem (see \eg\ Michalewicz~\cite{Mich2004}) or NP-hard spanning tree problems (see \eg\ Raidl and Julstrom~\cite{Raidl2003}).

Each of these representations induces a different neighborhood of a particular solution, and variation operators such as mutation and crossover have to be adjusted to the considered representation. Usually, such representations either lead directly to feasible solutions for the problem to be optimized or the search process is guided towards valid solutions by using some penalty functions. Here, the representation of possible solutions in combination with some suitable variation operators may be crucial for the success of the algorithm.

Recently, it has been proven for various combinatorial
optimization problems that they can be solved by
evolutionary algorithms in reasonable time using a suitable
representation together with mutation operators adjusted to
the given problem. Examples for this approach are the
single source shortest path
problem~\cite{ScharnowSorting2004}, all-pairs shortest path
problem \cite{dhk08}, multi-objective shortest path
problem~\cite{Horoba2009}, the travelling salesman
problem~\cite{theile09} and the knapsack
problem~\cite{Eremeev08}. The representations used in these
papers are different from the general encodings working
with binary strings as considered earlier in theoretical
works on the runtime behavior of evolutionary algorithms.
Instead, the chosen representations reflect some properties
of partial solutions of the problem at hand that allow to
obtain solutions that can be extended to optimal ones for
the considered problem. To obtain such partial solutions
the algorithms make use of certain diversity mechanisms
allowing the algorithms to proceed in a dynamic programming
way.

Note that the problem-solving capability of classical
genetic algorithms is sometimes explained using the
building block hypothesis~\cite{Goldberg}, which also
involves extension of partial solutions to the optimal
ones. A relation of the mentioned above EAs to dynamic
programming, however, allows to obtain more specific
results in terms of average optimization time.

Dynamic programming~(DP)~\cite{Bellman1962} is a well-known algorithmic technique that helps to tackle a wide range of problems. A general framework for dynamic programming has been considered by e.\,g. Woeginger~\cite{Woeginger2000} and Kl\"{o}tzler~\cite{Klotzler}. The technique allows to compute an optimal solution for the problem by extending partial solutions to an optimal one.

An important common feature of the evolutionary algorithms~\cite{dhk08,DoerrT09,Eremeev08, Horoba2009, ScharnowSorting2004, theile09} is that each of them is based on a suitable multi-objective formulation of the given problem. The schemes of these EAs and solution representations are different, however.

The algorithms proposed in~\cite{ScharnowSorting2004} and \cite{Eremeev08} are generalizations of the well-known (1+1)-EA (see e.g.~\cite{Beyer_howto}) to the multi-objective case and they are based on a different representation of solutions than the one used in our paper.

The ($\mu$ + 1)-EA in~\cite{theile09} employs a large population of individuals, where each individual encodes just one partial solution. In~\cite{dhk08,DoerrT09} it was shown that, for the all-pairs shortest path problem on an $n$-vertex graph, application of a suitable crossover operator can provably reduce the optimization time of the EA by a factor of almost $n^{3/4}$.

A special case of the DP-based evolutionary algorithm proposed in the present paper can be found, e.g. in~\cite{Horoba2009}. Both algorithms employ large populations of individuals where an individual encodes a partial solution. The outline of these algorithms is similar to that of the SEMO algorithm~\cite{LTZWD}.

Each gene in our problem representation defines one of the DP~transition mappings, and a composition of these mappings yields the DP~state represented by the individual. The proposed EA utilizes a mutation operator which is a special case of point mutation, where the gene subject to change is not chosen randomly as usual, but selected as the first gene which has never been mutated so far (see Section~\ref{subsec:modules} for details and links to the biological systems).

The goal of the paper is to relate the above mentioned multi-objective evolutionary approaches to dynamic programming and give a general setup for evolutionary algorithms that are provably able to solve problems having a dynamic programming formulation. In particular, we show that in many cases a problem that can be solved by dynamic programming in time $T$ has an evolutionary algorithm which solves it in expected time $\bigo{T \cdot n \cdot \log{(|DP|)}}$ with $n$ being the number of phases and $|DP|$ being the number of states produced at the completion of dynamic programming.

The obtained results are not aimed at the development of faster solution methods for the combinatorial optimization problems (to construct an EA in our framework, one has to know enough about the problem so that the traditional DP algorithm could be applied and this algorithm would be more efficient). Instead, we aim at characterizing the area where evolutionary algorithms can work efficiently and study the conditions that ensure this. To put it informally, our results imply that a class of problems that is easy for the DP algorithm is also easy for a suitable EA for most of the reasonable meanings of the term ``easy'' (solvable in polynomial or pseudo-polynomial running time or admitting FPTAS).

\subsection{Organization}

The rest of the paper is organized as follows. In Section~\ref{sec:dynamic_programming}, we introduce a general dynamic programming formulation and the kind of problems that we want to tackle. This dynamic programming approach is transferred into an evolutionary algorithm framework in Section~\ref{sec:evolutionary_algorithms}. Here we also show how to obtain evolutionary algorithms carrying out dynamic programming for some well-known combinatorial optimization problems. In Section~\ref{sec:FPRAS}, we consider a wide class of the DP-benevolent problems which are known to have fully polynomial-time approximation schemes based on dynamic programming~\cite{Woeginger2000}. We show that for the problems of this class there exists a fully polynomial-time randomized approximation scheme based on an evolutionary algorithm. Finally, we finish with some conclusions.

The main results of Sections~\ref{sec:dynamic_programming} and~\ref{sec:evolutionary_algorithms} originally were sketched in our extended abstract~\cite{dehnt}, while the main result of Section~\ref{sec:FPRAS} was published in Russian in~\cite{Eremeev10}. Additionally to refined presentation of results~\cite{dehnt, Eremeev10}, the present paper contains a DP-based EA with a strengthened runtime bound for the case of DP algorithm with homogeneous transition functions (applicable e.g. to the shortest path problems).


\section{Dynamic Programming}
\label{sec:dynamic_programming}

Dynamic programming is a general design paradigm for
algorithms. The basic idea is to divide a problem into
subproblems of the same type, and to construct a solution
for the whole problem using the solutions for the
subproblems. Dynamic programming has been proven to be
effective for many single-objective as well as
multi-objective optimization problems. It is even the most
efficient approach known for solution of some problems in
scheduling~\cite{PottsKovalyov,Woeginger2000},
bioinformatics~\cite{CHSR2010}, routing (see
e.g.~\cite{CLRC01}, Chapters~24, 25) and other areas.

In this section, we will assume that an original optimization problem~$\Pi$ (single-objective or multi-objective) may be transformed into a multi-objective optimization problem~$P$ of a special type. The general scheme of dynamic programming will be presented and studied here in terms of the problem~$P$. Several examples of a transformation from $\Pi$ to~$P$ are provided at the end of the section.

\subsection{Multi-Objective Optimization Problem}

Let us consider a multi-objective optimization problem~$P$ which will be well suited for application of the DP algorithm in some sense, as shown below. Suppose, there are $d \in \Nats$ objectives that have to be optimized in~$P$. An instance of problem~$P$ is defined by a quadruple $(d,g,\Space,{\mathcal D})$. Here $g \colon \Space \to (\Reals\cup \{\infty\})^{d}$ is called the \emph{objective function}, $\Space$ is called the \emph{search space}, and $g(\Space) \subseteq (\Reals\cup \{\infty\})^{d}$ is the \emph{objective space}. ${\mathcal D} \subseteq {\mathcal S}$ is a set of feasible solutions.

We introduce the following partial order to define the goal in multi-objective optimization formally. Throughout this paper, $\preceq$ denotes \emph{Pareto dominance} where
$$
    (y_{1},\dots,y_{d}) \preceq (y'_{1},\dots,y'_{d})
$$
iff $y_{i} \geq y'_{i}$ for all $i$ for minimization criteria~$g_i$ and $y_{i} \leq y'_{i}$ for maximization criteria~$g_i$. In the following, we use the notation $y' \prec y$ as an abbreviation for $y' \preceq y$ and $y \not\preceq y'$. The \emph{Pareto front} is the subset of $g({\mathcal D})$ that consists of all maximal elements of $g({\mathcal D})$ with respect to $\preceq$. The goal is to determine a \emph{Pareto-optimal set}, that is, a minimal by inclusion subset of feasible solutions~${\mathcal D}$ that is mapped on the Pareto front.

\subsection{Framework for Dynamic Programs}
\label{sub:framework_for_dynamic_programs}

Consider a DP algorithm for a problem~$P$, working through a number of iterations called {\em phases}. In each phase the DP algorithm constructs and stores some {\it states} belonging to~$\Space$. By saying that DP algorithm computes a Pareto-optimal set for the problem~$P$ we mean that after completion of the DP algorithm, the set of all DP states produced at the final phase is a Pareto-optimal set for~$P$.

Application of the DP approach to many multi-objective and single-objective optimization problems can be viewed as a transformation of a given problem~$\Pi$ to some problem~$P$: a DP algorithm is applied to compute a Pareto-optimal set for~$P$ and this set is efficiently transformed into a solution to the given single- or multi-objective problem.

In what follows, we consider only those DP algorithms where the states of the current phase are computed by means of {\em transition functions}, each such function depending on the input parameters of problem~$P$ and taking as an argument some state produced at the previous phase.

Let us start the formal definition of the DP algorithm from a simplified version. Suppose that the simplified DP algorithm works in $n$ phases, such that in the $i$-th phase a set $\mathcal{S}_i \subseteq \Space$ of states is created. We use $n$ finite sets $\mathcal{F}_{i}$ of state transition functions $F \colon \Space \to \Space'$ to describe the DP algorithm. Here $\Space'$ is an extension of space $\Space$. A mapping $F$ can produce elements $F(S) \in \Space' \backslash \Space$ that do not belong to a search space. To discard such elements at phase~$i$, $i=1,\dots,n$, a {\em consistency function} $H_i$ is used, $H_i \colon \Space' \to \Reals$, such that $S \in \Space$ iff $H_i(S) \leq 0$. We assume that the number~$n$, the functions~$H_i$ and the sets of functions~$\mathcal{F}_{i}$ depend on the input instance of problem~$P$.

The simplified DP algorithm proceeds as follows. In the initialization phase, the state space $\mathcal{S}_{0}$ is initialized with a finite subset of $\Space$. In the $i$-th phase, the state space $\mathcal{S}_{i}$ is computed using the state space $\mathcal{S}_{i-1}$ according to
\begin{equation}
    \label{eqn:ext}
    \mathcal{S}_{i} = \{F(S) \mid S \in \mathcal{S}_{i-1} \wedge F \in \mathcal{F}_{i} \wedge H_i(F(S)) \leq 0\}.
\end{equation}
In the process, the consistency functions~$H_i$ serve to keep the infeasible elements emerging in phase~$i$ from being included into the current state space~$\mathcal{S}_{i}$. (Note that after completion of phase~$n$ of the simplified DP algorithm, the set~$\mathcal{S}_{n}$ may contain some states whose objective values are Pareto-dominated by those of other states from~$\mathcal{S}_{n}$.)

To delete the states with Pareto-dominated objective values and to improve the runtime of the simplified DP algorithm defined by~(\ref{eqn:ext}), most of the practical DP algorithms utilize the Bellman principle (see \eg\ \cite{Bellman1962}) or its variations so as to dismiss unpromising states without affecting the optimality of the final set of solutions. A formulation of the Bellman principle in terms of recurrence~(\ref{eqn:ext}) for the single-objective problems can be found in \ref{app:bellman}. Sufficient conditions for application of the Bellman principle in the single-objective case were formulated in~\cite{Mitten}. In the multi-objective case the Bellman principle is not used, but the unpromising states may be excluded by means of an appropriate dominance relation on the set of states. Originally such dominance relations were introduced by R.~Kl\"{o}tzler~\cite{Klotzler}. In this paper, we employ a similar approach, motivated by~\cite{Woeginger2000}.

Let us consider a partial quasi-order (\ie a reflexive and transitive relation)~$\preceq_{\mathrm{dom}}$ defined on $\Space$ so that $S \preceq_{\mathrm{dom}} S'$ iff $g(S) \preceq g(S')$. We will say that state~$S$ is \emph{dominated} by state~$S'$ iff $S \preceq_{\mathrm{dom}} S'$. If $S\in {\mathcal T} \subseteq \Space$ is such that no $S'\in {\mathcal T}$ exists satisfying $S \preceq_{\mathrm{dom}} S'$, then~$S$ will be called {\em non-dominated} in~${\mathcal T}$.

As we will see, under the following two conditions the relation $\preceq_{\mathrm{dom}}$ is helpful to dismiss unpromising states in the DP algorithm.

The first condition C.1 guarantees that the dominance relation between two states transfers from one round to the next:

Condition C.1.
    For any $S,S' \in \mathcal{S}_{i-1}, i=1,\dots,n$, if $S \preceq_{\mathrm{dom}} S'$ then $F(S) \preceq_{\mathrm{dom}} F(S')$ for all $F \in \mathcal{F}_{i}$.

The second condition C.2 expresses that infeasible states cannot dominate feasible states:

Condition C.2.
    For any $S,S' \in \mathcal{S'}$, if $S \preceq_{\mathrm{dom}} S'$ and ${H_i(S) \leq 0}$ then $H_i(S') \leq 0$.

Consider a subset $\mathcal{S}_{i}$ of $\Space$. We call $\mathcal{T}_{i} \subseteq \mathcal{S}_{i}$ a \emph{dominating subset} of $\mathcal{S}_{i}$ with respect to $\preceq_{\mathrm{dom}}$ iff for any state $S \in \mathcal{S}_{i}$ there is a state $S' \in \mathcal{T}_{i}$ with $S \preceq_{\mathrm{dom}} S'$. Let us use the notation $M(\mathcal{S}_{i}, \preceq_{\mathrm{dom}})$ to denote the set of all dominating subsets of $\mathcal{S}_{i}$ which are minimal by inclusion.

The following proposition indicates that under conditions C.1 and C.2 it  is sufficient to keep a dominating subset of states constructed in each phase~$i$, rather than the full subset $\mathcal{S}_{i}$.

\begin{proposition}
    \label{prn:ext_t}
    Suppose the simplified DP algorithm is defined by~(\ref{eqn:ext}),
    conditions C.1 and C.2 hold
    and the dominating sets $\mathcal{T}_{i}, \ i=1,\dots,n$ are computed
    so that $\mathcal{T}_{0} \in M(\mathcal{S}_{0},
    \preceq_{\mathrm{dom}})$,
    \begin{equation}\label{Tk_in_M}
        \mathcal{T}_{i} \in M(\{F(S) \mid S \in \mathcal{T}_{i-1} \wedge F \in \mathcal{F}_{i} \wedge H_i(F(S)) \leq 0\},
        \preceq_{\mathrm{dom}}).
    \end{equation}

    Then for any state $S^* \in {\mathcal S}_i, \ i=0,\dots,n,$
    there exists
    $S\in \mathcal{T}_{i}$ such that $S^* \preceq_{\mathrm{dom}} S$.
\end{proposition}

{\bf Proof.} The proof is by induction on $i$. For $i=0$ the statement holds by assumption $\mathcal{T}_0 \in M(\mathcal{S}_{0}, \preceq_{\mathrm{dom}})$.

By~(\ref{eqn:ext}), a state~$S^* \in {\mathcal S}_{i}$ can be expressed as $S^* = F^*(S')$, so that ${H_i(S^*)\le 0}$, $F^*\in \mathcal{F}_{i}$ and $S' \in {\mathcal S}_{i-1}$. But by induction hypothesis, there exists a state $S^{\diamond}\in \mathcal{T}_{i-1}$ such that $S' \preceq_{\mathrm{dom}} S^{\diamond}$. Now conditions C.1 and C.2 imply that $S^* = F^*(S') \preceq_{\mathrm{dom}} F^*(S^{\diamond})$  and $F^*(S^{\diamond}) \in \{F(S) \mid S \in \mathcal{T}_{i-1} \wedge F \in \mathcal{F}_{i} \wedge H_i(F(S)) \leq 0\}$. Hence, by~(\ref{Tk_in_M}) we conclude that there exists $S\in \mathcal{T}_{i}$ such that $S^*
\preceq_{\mathrm{dom}} F^*(S^{\diamond}) \preceq_{\mathrm{dom}} S$. \qed\\

In view of definition of~$\preceq_{\mathrm{dom}}$, if the conditions of Proposition~\ref{prn:ext_t} are satisfied and the Pareto front of $g$ is contained in $g(\mathcal{S}_{n})$, then this Pareto front is also contained in $g(\mathcal{T}_{n})$.

\begin{proposition}\label{prop:note}
If the conditions of Proposition~\ref{prn:ext_t} are satisfied, then the size of each $\mathcal{T}_i, \ i=0,\dots,n,$ is uniquely determined.
\end{proposition}

Indeed, consider the set of maximal elements of ${\mathcal S}_i$ with respect to $\preceq_{\mathrm{dom}}$. Define the equivalence classes of this set with respect to the equivalence relation $x \equiv y$ iff $x \preceq_{\mathrm{dom}} y$ and $y \preceq_{\mathrm{dom}} x$. The size of a minimal subset~$M$ of the set of maximal elements, which dominates all elements of ${\mathcal S}_i$, is unique since such~$M$ contains one representative element from each equivalence class. \qed

A computation satisfying~(\ref{Tk_in_M}) can be expressed in an algorithmic form as presented in Algorithm~\ref{alg:dp}. It is easy to see that when a subset~$\mathcal{T}_{i}$ is completed in Lines 8-13, condition~(\ref{Tk_in_M}) holds.

\begin{algorithm}[t]
    \caption{Dynamic Program for~$P$}
    \begin{algorithmic}[1]
        \STATE $\mathcal{T}_{0} \leftarrow \emptyset$
        \FOR{$S \in \mathcal{S}_{0}$}
            \IF{$\nexists S' \in \mathcal{T}_{0} \colon S \preceq_{\mathrm{dom}} S'$}
                \STATE $\mathcal{T}_{0} \leftarrow (\mathcal{T}_{0} \setminus \{S' \in \mathcal{T}_{0} \mid S' \prec_{\mathrm{dom}} S\}) \cup \{S\}$
            \ENDIF
        \ENDFOR
        \FOR{$i=1$ to $n$}
            \STATE $\mathcal{T}_{i} \leftarrow \emptyset$
            \FOR{$S \in \mathcal{T}_{i-1}$ and $F \in \mathcal{F}_{i}$}
                \IF{$H_i(F(S)) \leq 0$ and $\nexists S' \in \mathcal{T}_{i} \colon F(S) \preceq_{\mathrm{dom}} S'$}
                    \STATE $\mathcal{T}_{i} \leftarrow (\mathcal{T}_{i} \setminus \{S' \in \mathcal{T}_{i} \mid S' \prec_{\mathrm{dom}} F(S)\}) \cup \{F(S)\}$
                \ENDIF
            \ENDFOR
        \ENDFOR
        \RETURN $\mathcal{T}_{n}$
    \end{algorithmic}
    \label{alg:dp}
\end{algorithm}

The runtime of a DP algorithm depends on the computation times for the state transition functions $F \in \mathcal{F}_{i}$, for the consistency functions $H_i$, for checking the dominance and manipulations with the sets of states. Let $\theta_{F}$ be an upper bound on computation time for a transition function~$F$ and let $\theta_{\mathcal{H}}$ be an upper bound for computation time of any function $H_i, \ i=1,\dots, n$. Sometimes it will be appropriate to use the average computation time for the state transition functions at phase~$i, \ i=1,\dots,n$: $\theta_{\mathcal{F}_i}=\sum_{F \in \mathcal{F}_{i}} \theta_{F} / |\mathcal{F}_{i}|$.

In Algorithm~\ref{alg:dp}, verification of condition
\begin{equation}\label{verify0}
\nexists S' \in \mathcal{T}_{i} \colon F(S)
\preceq_{\mathrm{dom}} S'
\end{equation}
in Line~10 and execution of Line~11 may be implemented using similar problem-specific data structures. To take this into account, we will denote by~$\theta_{\preceq}$ an upper bound applicable both for the time to verify~(\ref{verify0}) and for the time to execute Line~11.

The body (lines 10--12) of the main loop (Lines 7--14) in Algorithm~\ref{alg:dp} is executed $\sum_{i=1}^{n}{|\mathcal{T}_{i-1}| \cdot |\mathcal{F}_{i}|}$ times.

To simplify the subsequent analysis let us assume that in the if-statement at line~10, the condition~(\ref{verify0}) is always checked. We denote the computation time for initializing $\mathcal{T}_{0}$ with $\theta_{\mathrm{ini}}$ (Lines 1--6) and the computation time for presenting the result with $\theta_{\mathrm{out}}$ (Line 15), which leads to an overall runtime
\begin{equation}
    \label{eqn:DPruntime}
    \bigo{\theta_{\mathrm{ini}} + \sum_{i=1}^{n}{|\mathcal{F}_{i}| \cdot |\mathcal{T}_{i-1}| \cdot (\theta_{\mathcal{F}_i} + \theta_{\mathcal{H}} + \theta_{\preceq})} + \theta_{\mathrm{out}}}.
\end{equation}
In many applications of the DP, the computation time for the state transition functions and the consistency functions are constant. Besides that, the partial quasi-order~$\succeq_{\mathrm{dom}}$ is often just a product of linear orders and it is sufficient to allocate one element in a memory array to store one (best found) element for each of the linear orders. This data structure usually allows to verify~(\ref{verify0}) and to execute Line~11 in constant time (see the examples in Subsection~\ref{sec:examples}). In the cases mentioned above, the values $\theta_{F}$, $\theta_{\mathcal{H}}$ and $\theta_{\preceq}$ can be chosen equal to the corresponding computation times and the overall DP algorithm runtime in~(\ref{eqn:DPruntime}) can be expressed with symbol $\Theta(\cdot)$ instead of $\bigo{\cdot}$.

Note that the runtime of  the DP algorithm is polynomially bounded in the input length of problem~$P$ if $\theta_{\mathrm{ini}}$, $n$, $\theta_{\mathcal{F}_i}$, $\theta_{\mathcal{H}}$, $\theta_{\preceq}$, $\theta_{\mathrm{out}}$, as well as $|\mathcal{T}_{i}|$ and $|\mathcal{F}_{i+1}|$ for $i=0,\dots,n-1$,  are polynomially bounded in the input length. Here and below, we say that a value (e.g. the running time) is polynomially bounded in the input length, meaning that there exists a polynomial function of the input length, which bounds the value from above.

\subsection{Applications of the general DP scheme}
\label{sec:DP_examples}

In this subsection, we point out how the general DP framework presented above is applied to some classical combinatorial optimization problems. The approach followed here is to describe the appropriate problem~$P$ and the components of a dynamic programming algorithm for the solution of a specific problem~$\Pi$. Most of the following examples have been inspired by the previous works~\cite{dhk08,ScharnowSorting2004,theile09}. Note that $\preceq_{\mathrm{dom}}$ will be a product of linear orders in each of these examples. In what follows $\mbox{id}$ denotes the identical mapping.

\paragraph{Traveling Salesman Problem}

Let us first consider the traveling salesman problem (TSP) as a prominent NP-hard example. The input for the TSP consists of a complete graph~${\mathcal G}=(V,E)$ with a set of nodes $V=\{1,2,\dots,n\}$ and non-nega\-tive edge weights $w \colon E \to \Reals_{0}^{+}$. It is required to find a permutation of all nodes $(v_1,\dots,v_n)$, such that the TSP tour length $\sum_{i=2}^{n}{w(v_{i-1},v_{i})} + w(v_{n},v_{1})$ is minimized. Without loss of generality we can assume that $v_1=1$, that is, the TSP tour starts in the fixed vertex $1$.

The search  space $\Space$ for problem~$P$ corresponding to the dynamic programming algorithm of Held and Karp~\cite{hk} consists of all paths~$S=(v_1,\dots,v_i), v_1=1$ of $i=1,\dots,n$ nodes. $\Space'$ is the extended search space of all sequences of nodes up to length~$n$ (the same node may occur more than once). Given $M\subseteq V\backslash \{1\}$ and $k\in M$, let $\pi(k,M)$ denote the set of all paths of $|M|+1$ vertices starting in vertex $1$ then running over all nodes from $M$ and ending in vertex~$k$. Let the vector objective function~$g:\Space \to (\Reals\cup \{\infty\})^{d}$, $d=(n-1)2^{(n-1)}$ have components $g_{kM}(S)$ for all $M\subseteq V\backslash \{1\},$ $k\in M$, equal to the length of path $S$ iff $S \in \pi(k,M)$. For all other $S \not \in \pi(k,M)$ assume $g_{kM}(S)=\infty$. The set of feasible solutions is ${\mathcal D}=\cup_{k=2}^n \pi(k,V\backslash \{1\})$, since in the TSP we seek a tour of length~$n$.

${\cal S}_0$ consists of a single element~$v_1$. The set~${\cal F}_i$ for all $i$ consists of~$n-1$ functions $F_v \colon \Space \to \Space'$ that add vertex $v\in V \backslash \{1\}$ to the end of the given path. For invalid states~$S\in \Space'$, which are characterized by not being Hamiltonian paths on their vertex sets, the mapping $H_i(S)$ computes $1$ and $0$ otherwise.

In view of the definition of objective~$g$, the dominance relation is formulated as follows. $S \preceq_{\mathrm{dom}} S'$ if and only if $S$ and $S'$ are Hamiltonian paths on the same ground set with the same end vertex $k$ and path $S'$ is not longer than $S$. States from different sets $\pi(k,M)$ are not comparable. Conditions C.1 and C.2 are verified straightforwardly.

Substituting these components into Algorithm~\ref{alg:dp},  we get almost the whole well-known dynamic programming algorithm of Held and Karp~\cite{hk}, except for the last step where the optimal tour is constructed from the optimal Hamiltonian paths.

Algorithm~\ref{alg:dp} initializes the states of the dynamic program with paths $(1,v)$ for all $v\in V \setminus \{1\}$. In each subsequent iteration~$i$, the algorithm takes each partial solution~$S$ obtained in the preceding iteration and checks for every application of the state transition function $F(S)$ with $F\in\mathcal{F}_i$ whether $H_i(F(S))$ is a feasible partial solution that is non-dominated in $\mathcal{T}_i$. If so, then $F(S)$ is added to the set $\mathcal{T}_i$ of new partial solutions by replacing dominated partial solutions $S'$ defined on the same ground set with the same end vertex of the Hamiltonian path.

What remains to do after completion of the DP algorithm with Pareto-optimal set is to output the Pareto-optimal solution minimizing the criterion $g_{k,V\backslash \{1\}}(S)+w(k,1), k\in V\backslash \{1\}$, which is now easy to find. Here using appropriate data structures one gets $\theta_{F}=\Theta(1)$, $\theta_{\mathcal{H}}=\Theta(1)$, $\theta_{\preceq}=\Theta(1)$ and $|{\mathcal T}_i|=i{n-1 \choose i}$, $|{\mathcal F}_{i}|=n-1$ for all $i=1,\dots,n$, thus the observation following~(\ref{eqn:DPruntime}) leads to the time complexity bound $\Theta(n^2 2^n)$.

\paragraph{Knapsack Problem}

Another well-known NP-hard combinatorial optimization problem that can be solved by dynamic programming is the knapsack problem. The input for the knapsack problem consists of $n$ items where each item $i$ has an associated integer weight $w_i>0$ and profit $p_i>0$, $1 \leq i \leq n$. Additionally a weight bound $W$ is given. The goal is to determine an item selection $K \subseteq \{1,\dots,n\}$ that maximizes the profit $\sum_{i \in K}{p_{i}}$, subject to the condition $\sum_{i \in K}{w_{i}} \leq W$.

We fit the problem into the above framework assuming that each state $S=(s_1,s_2)\in {\cal S}_i, \ i=1,\dots,n$, encodes a partial solution for the first~$i$ items, where coordinate~$s_1$ stands for the weight of a partial solution and~$s_2$ is its profit. The initial set ${\mathcal S}_0$ consists of a single element $(0,0)$ encoding a selection of no items.

The pseudo-Boolean vector function $g \colon \Space \to \Reals^W$ defines $W$ criteria
\begin{equation}
g_w(S) := \left\{
  \begin{array}{ll}
  s_2 &\mbox{ if } s_1 =w\\
   0 & \mbox{otherwise}
  \end{array} \right. , \quad w=0,\dots,W,
\end{equation}
that have to be maximized. This implies the dominance relation $\preceq_{dom}$ such that $S \preceq_{\mathrm{dom}} S'$ iff $s_1 = s'_1$ and $s_2 \le s'_2$, where $S=(s_1,s_2), \ S'=(s'_1,s'_2)$.

The set~${\mathcal F}_i$ consists of two functions: $\mbox{id}$ and $ F_i(s_1,s_2) = (s_1+w_i, s_2+p_i).$ Here $F_i$ corresponds to adding the $i$-th item to the partial solution, and $\mbox{id}$ corresponds to skipping this item. A new state $S=(s_1,s_2)$ is accepted if it does not violate the weight limit, \ie $H_i(S) \leq 0$, where $H_i(S) = s_1 - W$.

The conditions C.1 and C.2 are straightforwardly verified. To obtain an optimal solution for the knapsack problem it suffices to select the Pareto-optimal state with a maximal component~$s_2$ from $\mathcal{S}_{n}$.

To reduce the comparison time $\theta_{\preceq}$ we can store the states of the DP in a $(W\times n)$-matrix. An element in row~$w, \ w=1,\dots,W$, and column~$i, \ i=1,\dots,n$, holds the best value~$s_2$ obtained so far on states $S=(s_1,s_2)\in{\mathcal T}_i$ with $s_1=w$. Then $\theta_{\preceq}$ is a constant and the worst-case runtime of the explained DP algorithm is $\bigo{n \cdot W}$ since $\sum_{i=1}^{n}{|\mathcal{T}_{i-1}|} \leq n W$.

\paragraph{Single Source Shortest Path Problem}
A classical problem that also fits into the DP framework is the single source shortest path problem (SSSP). Given an undirected connected graph $\mathcal{G} = (V,E)$, $|V|=n$ and positive edge weights $w \colon E \to \Reals^{+}$, the task is to find shortest paths from a selected {\em source} vertex $s \in V$ to all other vertices.

The  search space $\Space$ is a set of all paths in~$\mathcal{G}$
with an end-point~$s$. The set of feasible solutions~${\mathcal D}$
is equal to~$\Space$.

Since adding a vertex to a path may result in a sequence of vertices that do not constitute a path in~$\mathcal{G}$, we extend the search space to the set $\Space'$ of all sequences of vertices of length at most~$n$ with an end-point~$s$. The set~${\mathcal S}_0$ of initial solutions is just a single vertex~$s$. Now for all $i$, we define $\mathcal{F}_i := \{F_v \mid v \in V\} \cup \{\mbox{id}\}$, where $F_v \colon \Space \to \Space'$ is the mapping adding the vertex~$v$ to a sequence of vertices. $H_i(S)=-1$ if $S$ is a path in~$\mathcal{G}$ with an end-point~$s$, and $1$ if not.

Let the vector objective function~$g$ have $d=n$ components $g_{v}(S)$ for all $v\in V$, equal to the length of path $S$ iff $S$ connects $s$ to $v$, otherwise assume $g_{v}(S)=\infty$. This implies that $S \preceq_{\mathrm{dom}} S'$ if and only if the paths $S$ and $S'$ connect $s$ to the same vertex and $S'$ is not longer than $S$.

The resulting DP algorithm has $\theta_{F}=\Theta(1)$, $\theta_{\mathcal{H}}=\Theta(1)$, $\theta_{\preceq}=\Theta(1)$ and $|{\mathcal T}_{i-1}|=\Theta(n)$, $|{\mathcal F}_i|=\Theta(n)$ for all $i=1,\dots,n$, thus~(\ref{eqn:DPruntime}) gives the time complexity bound $O(n^3)$. The well-known Dijkstra's algorithm has $O(n^2)$ time bound, but in that algorithm only one transition mapping is applied in each phase (attaching the closest vertex to the set of already reached ones), and such a problem-specific DP scheme is not considered here.

\paragraph{All-Pairs Shortest Path Problem}
Finally, let us consider the all-pairs shortest path (APSP) problem, which has the same input as the SSSP, except that no source vertex is given, and the goal is to find for each pair $(u,v)$ of vertices a shortest path connecting them.

A basic observation is that sub-paths of shortest paths are shortest paths again. Hence a shortest path connecting $u$ and $v$ can be obtained from appending the edge $(x,v)$, where $x$ is a neighbor of $v$, to a shortest path from $u$ to $x$. This allows a very natural DP formulation as described for problem~$P$.

For the APSP, the search space $\Space$ naturally is the set of all paths in~$\mathcal{G}$, and the set~${\mathcal D}$ of feasible solutions consists of collections of paths, where for each pair of vertices there is one path connecting them.

We model paths via finite sequences of vertices, and do not allow cycles. Since adding a vertex to a path may create a sequence of vertices which does not correspond to a path in~$\mathcal{G}$, let us extend this search space to the set $\Space'$ of all sequences of vertices of length at most~$n$. The set~${\mathcal S}_0$ of initial solutions is the set of all paths of length $0$, that is, of all sequences consisting of a single vertex. Now for all $i$, we define $\mathcal{F}_i := \{F_v \mid v \in V\} \cup \{\mbox{id}\}$, where $F_v \colon \Space' \to \Space'$ is the mapping adding the vertex~$v$ to a sequence of vertices. To exclude invalid solutions, let us define $H_i(S)$ to be $-1$ if $S$ is a path in~$\mathcal{G}$, and $1$ if not.

It remains to define when one state dominates another. Let $\pi_{ij}$ denote the set of all paths starting in vertex $i$ and ending in vertex~$j$. Let the vector objective function~${g:\Space \to (\Reals\cup \{\infty\})^{d}}$, $d=n^2$ have components $g_{ij}(S)$ for all $i,j\in V$, equal to the length of path $S$ iff $S \in \pi_{ij}$. For all other $S \not \in \pi_{ij}$ assume $g_{ij}(S)=\infty$. This implies that  $S \preceq_{\mathrm{dom}} S'$ if and only if the paths $S$ and $S'$ connect the same two vertices and $S'$ is not longer than $S$.

Since the length of the path arising from extending an existing path by an edge depends monotonically on the length of the existing path, conditions C.1 and C.2 hold. So, in view of Proposition~\ref{prn:ext_t}, any set~${\mathcal T}_i$ contains a path for each pair of vertices (and only one such path). Thus, ${\mathcal T}_n$ is a subset of~${\mathcal D}$ and contains a shortest path for any pair of vertices.

The resulting algorithm following the dynamic programming approach now does the following. It starts with all paths of length zero as solution set~${\mathcal S}_0$. It then repeats $n$ times the following. For each path in the solution set and each vertex, it appends the vertex to the path. If the resulting path dominates an existing solution with the same end vertices, it replaces the latter. Here $\theta_{F}=\Theta(1)$, $\theta_{\mathcal{H}}=\Theta(1)$, $\theta_{\preceq}=\Theta(1)$ and $|{\mathcal T}_i|=O(n^2)$, $|{\mathcal F}_i|=O(n)$ for all $i=1,\dots,n$, thus~(\ref{eqn:DPruntime}) gives the time complexity bound $O(n^4)$. Note that the well-known Floyd-Warshall algorithm (see e.g.~\cite{CLRC01}, Chapter~25) has $O(n^3)$ time bound, but in that algorithm each transition mapping combines two states (paths), and such an option is not considered in this paper.


\section{Evolutionary Algorithms}
\label{sec:evolutionary_algorithms}

In the following, we show how results of dynamic programming can be attained by evolutionary algorithms. To this aim, we state a general formulation of such an evolutionary algorithm and then describe how the different components have to be designed.

\subsection{Framework for Evolutionary Algorithms}

An evolutionary algorithm consists of different generic modules, which have to be made precise by the user to best fit to the problem. Experimental practice, but also some theoretical work (see \eg\ \cite{dhn2006, djeulerea07, myeuler07,n2004ec}), demonstrate that the right choice of representation, variation operators, and selection method is crucial for the success of such algorithms.

We assume again that an instance of problem~$P$ is given by a multi-objective function $g$ that has to be optimized. We consider simple evolutionary algorithms that consist of the following components.

We use $\Space'_{\mathrm{EA}} := \{0,\dots,n\} \times \Space'$ as the {\em phenotype space} and call its elements {\em individuals}. The algorithm (see Algorithm~\ref{alg:ea}) starts with an initial population of individuals $\mathcal{P}_{0}$. During the optimization the evolutionary algorithm uses a selection operator $\mbox{sel}(\cdot)$ and a mutation operator $\mbox{mut}(\cdot)$ to create new individuals. The $d$-dimensional objective function together with a partial order $\preceq$ on $\Reals^{d}$ induce a partial quasi-order $\preceq_{\mathrm{EA}}$ on the phenotype space, which guides the search. After the termination of the EA, an output function $\mbox{out}_{\mathrm{EA}}(\cdot)$ is utilized to map the individuals in the last population to search points from the DP search space.

\begin{algorithm}[t]
    \caption{Evolutionary Algorithm for $P$}
    \begin{algorithmic}[1]
        \STATE $\mathcal{P} \leftarrow \emptyset$
        \FOR{$I \in \mathcal{P}_{0}$}
            \IF{$\nexists I' \in \mathcal{P} \colon I \prec_{\mathrm{EA}} I'$}
                \STATE $\mathcal{P} \leftarrow (\mathcal{P} \setminus \{I' \in \mathcal{P} \mid I' \prec_{\mathrm{EA}} I\}) \cup \{I\}$
            \ENDIF
        \ENDFOR
        \LOOP
            \STATE $I \leftarrow \mbox{mut}(\mbox{sel}(\mathcal{P}))$
            \IF{$\nexists I' \in \mathcal{P} \colon I \prec_{\mathrm{EA}} I'$}
                \STATE $\mathcal{P} \leftarrow (\mathcal{P} \setminus \{I' \in \mathcal{P} \mid I' \prec_{\mathrm{EA}} I\}) \cup \{I\}$
            \ENDIF
        \ENDLOOP
        \RETURN $\{\mbox{out}_{\mathrm{EA}}(I) \mid I=(i,S) \in {\mathcal P}, S \in {\mathcal D}\}$
    \end{algorithmic}
    \label{alg:ea}
\end{algorithm}

\subsection{Defining the Modules} \label{subsec:modules}

We now consider how the different modules of the evolutionary algorithm have to be implemented so that it can carry out dynamic programming. To do this, we relate the modules to the different components of a DP algorithm. Consider a problem~$P$ given by a set of feasible solutions~${\mathcal D}$ and a multi-objective function $g$ that can be solved by a dynamic programming approach. The EA works with the following setting.

The initial population is $\mathcal{P}_{0} = \{0\} \times \mathcal{S}_{0}$ where $\mathcal{S}_{0}$ is the initial state space of the DP algorithm. The selection operator $\mbox{sel}(\cdot)$ chooses an individual $I \in \mathcal{P}$ the following way. First it chooses~$i \le n-1$ uniformly from the set of phases which are represented in the current population \ie from the set $\{k : k \le n-1, \ \exists (k,S) \in {\cal P} \}$. After this, selection chooses~$I$ uniformly among the individuals of the form $(i,S)$ in the current population.

For an individual $(i,S)$, the mutation operator $\mbox{mut}(\cdot)$ chooses a state transition function $F \in \mathcal{F}_{i+1}$ uniformly at random and sets $\mbox{mut}((i,S)) = (i+1,F(S))$.

We incorporate a partial order $\preceq_{\mathrm{EA}}$ into the EA to guide the search. This relation is defined as follows:
\begin{equation}\label{Order_EA}
(i,S) \preceq_{\mathrm{EA}} (i',S') \Leftrightarrow
 (i=i' \ \mbox{and} \ S \preceq_{\mathrm{dom}} S')  \ \mbox{ or } H_i(S)
>0.
\end{equation}
Finally, we utilize the output function $\mbox{out}_{\mathrm{EA}}((i,S)) = S$ to remove the additional information at the end of a run of the EA. That is, we remove the information that was used to store the number of a certain round of the underlying dynamic program and transform an individual into a search point for the problem~$P$.

Note that the description of the Algorithm~\ref{alg:ea} does not employ the notion of the fitness function, although an appropriate multi-objective fitness function may be defined for compatibility with the standard EA terminology.

Finally, note that we do not discuss the solutions encoding in our EA because it is not essential for the analysis. However, it may be worth mentioning, when the biological analogy is considered. Here each of the genes $A_i, \ i=1,\dots,n$ would define the DP~transition mapping from a set~${\mathcal F}_i$, and a composition of these mappings would yield the DP~state represented by the individual. One of the possible options of each gene is ``undefined'', and the mutation operator modifies the first gene which is still ``undefined'' in the parent individual. A discussion of genetic mechanisms corresponding to the proposed mutation in a biological system is provided in \ref{app:mutation}.

\subsection{Runtime of the Evolutionary Algorithm}

Our goal is to show that the evolutionary algorithm solves the problem~$P$ efficiently if the dynamic programming approach does. To measure the time the evolutionary algorithm needs to compute a Pareto-optimal set for problem~$P$, one would analyze the expected number of fitness evaluations to come up with a Pareto-optimal set, when it is non-empty. This is also called the expected \emph{optimization time}, which is a common measure for analyzing the runtime behavior of evolutionary algorithms. The proposed EA does not use the multi-objective fitness function explicitly, but given enough memory, it may be implemented so that every individual constructed and evaluated in Lines~3~and~4 or in Lines~8-11 requires at most one evaluation of the objective function~$g$. Thus, we can define the optimization time for Algorithm~\ref{alg:ea} as $|\mathcal{S}_{0}|$ plus the number of iterations of the main loop (Lines~8-11) required to come up with a Pareto-optimal set. Analogous parameter of a
DP algorithm is the number of states computed during its execution.

The next theorem relates the expected optimization time of the EA to the number of states computed during the execution of the corresponding DP algorithm. In what follows it will be convenient to denote the cardinality of the set of states produced after completion of the DP algorithm by~$|DP|$, \ie $|DP|:=\sum_{i=0}^{n}{|\mathcal{T}_i|}$. Note that $|DP|$ is a well-defined value since the sizes $|\mathcal{T}_i|$ are unique according to Proposition~\ref{prop:note}.

\begin{theorem}
    \label{thm:relation_between_EAs_and_DPs}
    Let a DP be defined as in Algorithm~\ref{alg:dp} and an EA defined as in Algorithm~\ref{alg:ea}
    with $\preceq_{\mathrm{EA}}$ relation defined by~(\ref{Order_EA}).
    Then the number of states computed during the execution of the DP algorithm is
    $|\mathcal{S}_{0}|+\sum_{i=1}^{n}{|\mathcal{F}_{i}| \cdot |\mathcal{T}_{i-1}|}$,
    and the EA has an expected optimization time of
$$
    \bigo{|\mathcal{S}_{0}| + n \cdot \log{|DP|} \cdot
    \sum_{i=0}^{n-1}|\mathcal{T}_{i}| \cdot |\mathcal{F}_{i+1}|}.
$$
\end{theorem}

{\bf Proof.} Estimation of the number of states computed during the execution of the DP algorithm is straightforward.

Assume that the optimization process works in stages $1 \le i \le n$, whereas stage $i+1$ starts after the stage $i$ has been finished. We define that a stage~$i$ finishes when for every state $S \in \mathcal{T}_i$ there exists an individual $(i,S') \in  \mathcal{P}$ with~$S'$ dominating $S$. Here and below in this proof, by $\mathcal{T}_i, \ i=0,\dots,n$, we denote the corresponding sets computed in Algorithm~\ref{alg:dp}. Note that after completion of a stage~$i$, the subset of individuals of a form $(i,S)$ in population ${\mathcal P}$ does not change in the subsequent iterations of the EA. Let $\mathcal{T}'_i$ denote the set of states of these individuals after completion of stage~$i, \ i=0,\dots,n$. By the definition of Algorithm~\ref{alg:ea}, the sequence $\mathcal{T}'_i, \ i=0,\dots,n$ satisfies~(\ref{Tk_in_M}), and therefore $|\mathcal{T}'_i|=|\mathcal{T}_i|, \ i=0,\dots,n$ in view of Proposition~\ref{prop:note}.

Let $\xi_i$ be the random variable denoting the number of iterations since stage~$i-1$ is finished, until stage $i$ is completed. Then the expected optimization time is given by $|\mathcal{S}_{0}|+E[\xi]$ with $\xi = \xi_1 + \dots +\xi_n$.

Any state~$S \in \mathcal{T}_{i+1}$ is computed in Algorithm~\ref{alg:dp} by means of some function $\tilde{F} \in {\mathcal F}_{i+1}$, when it is applied to some state $\tilde{S} \in \mathcal{T}_{i}$. Thus, in stage~$i+1$ of the EA during mutation the same transition function $\tilde{F}$ may be applied to some individual $I'=(i,S')$, such that $\tilde{S} \preceq_{\mathrm{dom}} S'$. After this mutation, in view of conditions C.1 and C.2, the population $\mathcal{P}$ will contain an individual $I''=(i+1,S'')$ with $S''$ such that $S \preceq_{\mathrm{dom}} \tilde{F}(S') \preceq_{\mathrm{dom}} S''$.

Consider any iteration of the EA at stage~$i+1$. Let~$t$ denote the number of such states from $\mathcal{T}_{i+1}$ that are already dominated by a state of some individual in~${\mathcal P}$. Then there should be $|\mathcal{T}_{i+1}|-t$ new individuals of the form $(i+1,S)$ to be added into~${\mathcal P}$ to complete stage~$i+1$ (recall that $|\mathcal{T'}_{i+1}|=|\mathcal{T}_{i+1}|$). The probability to produce an individual~$(i,S')$ where $S'$ dominates a previously non-dominated state from~$\mathcal{T}_{i+1}$ is no less than $(|\mathcal{T}_{i+1}| - t) / (n |\mathcal{T}_{i}| \cdot |\mathcal{F}_{i+1}|)$ with an expected waiting time of at most $(n |\mathcal{T}_{i}| \cdot |\mathcal{F}_{i+1}|) / (|\mathcal{T}_{i+1}| - t)$ for this geometrically distributed variable. The expected waiting time to finish stage $i+1$ is thus bounded by
    $$
        E[\xi_{i+1}] \le \sum_{t=1}^{|\mathcal{T}_{i+1}|}{\frac{n |\mathcal{T}_{i}| \cdot |\mathcal{F}_{i+1}|}{t}} = n |\mathcal{T}_{i}| \cdot |\mathcal{F}_{i+1}| \cdot {\mathcal H}_{|\mathcal{T}_{i+1}|},
    $$
with ${\mathcal H}_k$ being the $k$-th harmonic number, ${\mathcal H}_{k} := \sum_{i=1}^{k}{\frac{1}{i}}$.

This leads to an overall expected number of iterations
$$
        E[\xi] \le \sum_{i=0}^{n-1}{n |\mathcal{T}_{i}| \cdot |\mathcal{F}_{i+1}| \cdot {\mathcal H}_{|\mathcal{T}_{i+1}|}}
        \le n (\ln{|DP|}+1) \cdot \sum_{i=0}^{n-1}|\mathcal{T}_{i}| \cdot |\mathcal{F}_{i+1}|.
$$
\qed\\

A similar inspection as in Subsection~\ref{sub:framework_for_dynamic_programs} reveals that the expected runtime of the EA is
$$
    \mathord{O}\mathord{\Big(\theta_{\mathrm{ini}} +
    n \log{|DP|} \cdot \\
    {}\sum_{i=0}^{n-1}\big(|\mathcal{F}_{i+1}| \cdot |\mathcal{T}_{i}| \cdot(
      \theta_{\mathcal{F}_i} + \theta_{\mathcal{H}} + \theta_{\preceq})\big) +
    \theta_{\mathrm{out}}\Big)},
$$
assuming the individuals of the population are stored in $n+1$ disjoint sets according to the first coordinate $i$.

As noted in Subsection~\ref{sub:framework_for_dynamic_programs}, if  the computation times for functions $F$, $H_i$ and dominance checking~(\ref{verify0}) as well as execution time for Line~11 in Algorithm~\ref{alg:dp} are constant, then $\theta_{F}$, $\theta_{\mathcal{H}}$ and $\theta_{\preceq}$ can be chosen {\em equal} to the corresponding computation times. In such cases a problem that is solved by dynamic programming Algorithm~\ref{alg:dp} in time~$T$, will be solved by the EA defined as in Algorithm~\ref{alg:ea} in expected time $\bigo{T n \log{|DP|}}$.

\subsection{Homogeneous transitions}

Some DP algorithms, like the ones for the APSP and SSSP problems, have a specific structure which may be exploited in the EA. In this subsection we consider the case of {\em homogeneous} transition functions where ${\mathcal F}_1\equiv\dots \equiv{\mathcal F}_n$ and $H_1 \equiv \dots \equiv H_n$. To simplify the notation in this case we will assume ${\mathcal F}_1 \equiv{\mathcal F}$ and $H_1(S)\equiv H(S)$. Additionally, we suppose that the identical mapping belongs to ${\mathcal F}$.

The formulated assumptions imply that once some state~$S$ is obtained in the DP algorithm, it will be copied from one phase to another, unless some other state will dominate it. Note also that it does not matter at what particular phase a state has been obtained -- the transition functions will produce the same images of this state. These observations motivate a modification of the partial order~$\preceq_{\mathrm{EA}}$, neglecting the phase number in comparison of individuals:
\begin{equation}\label{Order_EA_modified}
(i,S) \preceq_{\mathrm{EA}} (i',S') \Leftrightarrow S
\preceq_{\mathrm{dom}} S'  \ \mbox{ or } H_i(S) >0.
\end{equation}
In fact, now we can skip the index~$i$ in individuals~$(i,S)$ of the EA, so in this subsection the terms ``state'' and ``individual'' are synonyms and the phase number~$i$ is suppressed in the notation of individuals. As the following theorem shows, wider sets of comparable individuals in this special case allow to reduce the population size and thus improve the performance of the EA. Let us consider the {\em width} $W_{\mathrm{dom}}$ of partial order~$\preceq_{\mathrm{dom}}$, \ie the maximum size of a set of pairwise incomparable elements.

\begin{theorem}
    \label{thm:modified_EA_DP_relation}
    If the transition functions are homogeneous and $id \in {\mathcal
    F}$, then the EA defined as in Algorithm~\ref{alg:ea}
    with the modified $\preceq_{\mathrm{EA}}$ relation~(\ref{Order_EA_modified}) has an expected optimization time of
    $
    \bigo{|\mathcal{S}_{0}|+W_{\mathrm{dom}} \log({W_{\mathrm{dom}}}) \cdot n |\mathcal{F}|}.
    $
\end{theorem}

{\bf Proof.} The analysis is similar to the proof of Theorem~\ref{thm:relation_between_EAs_and_DPs}. Note that now the size of population~$\mathcal{P}$ does not exceed~$W_{\mathrm{dom}}$. We assume that $|\mathcal{P}| = W_{\mathrm{dom}}$ right from the start.

Let $\mathcal{T}_i$ be the same as in phase~$i$ of the DP algorithm, $i=0,\dots,n$. Suppose again that the optimization process works in stages $1 \le i \le n$, whereas stage $i$ is assumed to be finished when for every $S \in \mathcal{T}_i$, the population $\mathcal{P}$ contains an individual $S'$ such that $S \preceq_{\mathrm{dom}} S'$.

Let $\xi_i$ be the number of iterations since stage~$i-1$ is finished, until stage $i$ is completed. Then the expected optimization time is given by $|\mathcal{S}_{0}|+E[\xi]$ with $\xi = \xi_1 + \dots +\xi_n$.

Any state~$S \in \mathcal{T}_{i+1}$ is computed in the DP algorithm by means of some function $\tilde{F} \in {\mathcal F}_{i+1}$, when it is applied to some state $\tilde{S} \in \mathcal{T}_{i}$. Thus, in stage~$i+1$ of the EA during mutation the same transition function $\tilde{F}$ may be applied to some individual $I'=S'$, such that $\tilde{S} \preceq_{\mathrm{dom}} S'$. After this mutation, in view of conditions C.1 and C.2, the population $\mathcal{P}$ will contain an individual $I''=S''$ such that $S \preceq_{\mathrm{dom}} \tilde{F}(S') \preceq_{\mathrm{dom}} S''$.

The probability of such a mutation for a particular~$S \in \mathcal{T}_{i+1}$ is at least $1/(|{\mathcal F}_{i+1}| \cdot |\mathcal{P}|) \le 1/(|{\mathcal F}_{i+1}| \cdot W_{\mathrm{dom}})$. Let $t$ denote the number of states~$S \in \mathcal{T}_{i+1}$ that are already dominated at stage $i+1$. Then there are at least $|\mathcal{T}_{i+1}|-t$ possibilities to add a new individual, which dominates a previously non-dominated state from~$\mathcal{T}_{i+1}$. The probability for such a mutation is not less than $(|\mathcal{T}_{i+1}| - t) / (W_{\mathrm{dom}} \cdot |\mathcal{F}_{i+1}|)$ with an expected waiting time of at most $(W_{\mathrm{dom}} \cdot |\mathcal{F}_{i+1}|) / (|\mathcal{T}_{i+1}| - t)$ for this geometrically distributed variable. The expected waiting time to finish stage $i+1$ is thus $E[\xi_{i+1}] \le W_{\mathrm{dom}} \cdot |\mathcal{F}_{i+1}| \cdot {\mathcal H}_{|\mathcal{T}_{i+1}|}$. But $|{\mathcal T}_{i+1}| \le W_{\mathrm{dom}}$ because the states of ${\mathcal T}_{i+1}$ are pairwise incomparable according to Algorithm~\ref{alg:dp}. This leads to an overall expected number of iterations $E[\xi] \le W_{\mathrm{dom}} \cdot (\ln({W_{\mathrm{dom}}})+1) \cdot \sum_{i=0}^{n-1}|\mathcal{F}_{i+1}|.$ \qed\\


\subsection{Examples} \label{sec:examples}

Now, we point out how the framework presented in this section can be used to construct evolutionary algorithms using the examples from Section~\ref{sec:dynamic_programming}.

\paragraph{Traveling Salesman Problem}

Due to Theorem~\ref{thm:relation_between_EAs_and_DPs} the expected optimization time of the evolutionary algorithm based on the DP algorithm of Held and Karp presented in Section~\ref{sec:DP_examples} is $\bigo{n^4 2^n}$. This bound can be further improved to $\bigo{n^3 \cdot 2^n}$ for the EA proposed in \cite{theile09}.

\paragraph{Knapsack Problem}

Consider the DP algorithm presented in Section~\ref{sec:DP_examples}. The expected optimization time of the corresponding EA for the knapsack problem is $\bigo{n^{2} \cdot W \cdot \log{(n \cdot W)}}$ due to Theorem~\ref{thm:relation_between_EAs_and_DPs}.

\paragraph{Single Source Shortest Path Problem}

Application of Theorem~\ref{thm:relation_between_EAs_and_DPs} to the DP algorithm for SSSP problem from Section~\ref{sec:DP_examples} gives an expected optimization time of $\bigo{n^4 \log{(n)}}$ for Algorithm~\ref{alg:ea}.

The DP algorithm for SSSP problem has homogeneous transition functions with with $W_{\mathrm{dom}} = n$. Thus, the modified EA considered in Theorem~\ref{thm:modified_EA_DP_relation} has the expected optimization time $O(n^3 \log n)$. This bound can be further improved to $O(n^3)$ for the (1+1)-EA~\cite{ScharnowSorting2004}.

\paragraph{All-Pairs Shortest Path Problem}

Plugging the ideas of the DP algorithm for APSP problem presented in Section~\ref{sec:DP_examples} into the framework of Algorithm~\ref{alg:ea}, we obtain an EA with an expected optimization time of $\bigo{n^5 \log{(n)}}$ due to Theorem~\ref{thm:relation_between_EAs_and_DPs}.

It has been noted, however, that the DP algorithm for APSP has homogeneous transition functions, each set~${\mathcal F_i}$ contains the identical mapping. Here $W_{\mathrm{dom}} = n^2$, thus Theorem~\ref{thm:modified_EA_DP_relation} implies that the modified EA has the expected optimization time $O(n^4 \log n)$. This algorithm can be further improved to an EA with optimization time $\Theta(n^4)$ as has been shown in~\cite{dhk08}.


\section{Approximation Schemes}\label{sec:FPRAS}

In this section, we demonstrate that for many single-objective discrete optimization problems~$\Pi$ the above framework can be used to find feasible solutions with any desired precision. The supplementary multi-objective problem~$P$ will be formally introduced for compatibility with the previous sections, but it will not play a significant role here.

Throughout this section we assume that $\Pi$ is an NP-optimization problem~\cite{AP95}, ${\bf x}$ denotes the input data of an instance of~$\Pi$, $\mbox{\it Sol}_{\bf x}$ is the set of feasible solutions, ${m_{\bf x}:\mbox{\it Sol}_{\bf x} \to {{\mathbb N}_0}}$ is the objective function (here and below ${\mathbb N}_0$ denotes the set of non-negative integers). The optimal value of the objective function is $\mbox{\it OPT}({\bf x})=\max_{y \in \mbox{\it Sol}_{\bf x}} m_{\bf x}(y)$ if~$\Pi$ is a maximization problem, or $\mbox{\it OPT}({\bf x})=\min_{y \in \mbox{\it Sol}_{\bf x}} m_{\bf x}(y)$ in the case of minimization.
To simplify presentation in what follows we assume that~${\mbox{\it Sol}}_{\bf x} \ne \emptyset$.

To formulate the main result of this section let us start with two standard definitions~\cite{GJ79}.

A {\em $\rho$-approximation algorithm} for~$\Pi$ is an algorithm that for any instance~${\bf x}$ returns a feasible solution whose objective value at most $\rho$ times deviates from~$\mbox{\it OPT}({\bf x})$ (if the instance~${\bf x}$ is solvable). Such a solution is called {\em $\rho$-approximate}. A {\em fully polynomial time approximation scheme (FPTAS)} for a problem~$\Pi$ is a family of $(1+\varepsilon)$-approximation algorithms over all factors $\varepsilon>0$ with polynomially bounded running time in problem input size $|{\bf x}|$ and in $1/\varepsilon$.

In~\cite{Woeginger2000} G.~Woeginger proposed a very general FPTAS with an outline similar to the DP Algorithm~\ref{alg:dp}, except that the comparison of newly generated states to the former ones is modified so that the ``close'' states are not kept. This modified algorithm is denoted by DP$_{\Delta}$ in what follows (a detailed description of DP$_{\Delta}$ will be given in Subsection~\ref{subsec:fptas}).

The state space~${\mathcal S}$ and its subsets~${\mathcal T}_i$ computed in the DP Algorithm~\ref{alg:dp} may be exponential in problem input size, thus leading to an exponential running time of the DP algorithm (this holds e.g. for the Knapsack problem). The algorithm~DP$_{\Delta}$, however, iteratively thins out the state space of the dynamic program and substitutes the states that are ``close'' to each other by a single representative, thus bringing the size of the subsets~${\mathcal T}_i$ down to polynomial. This transformation is known as {\em trimming the state space} approach.

In~\cite{Woeginger2000}, a list of conditions is presented, that guarantee the existence of an FPTAS when there is an exact DP algorithm for a problem. If a problem~$\Pi$ satisfies these conditions, it is called {\em DP-benevolent}. This class, in particular, contains the knapsack problem and different scheduling problems, e.g. minimizing the total weighted job completion time on a constant number of parallel machines, minimizing weighted earliness-tardiness about a common non-restrictive due date on a single machine, minimizing the weighted number of tardy jobs etc. The definition of DP-benevolence is as follows.

The input data of~$\Pi$ has to be structured so that~${\bf x}$ consists of~$n$ vectors $X_1,\dots,X_n \in {{\mathbb N}_0^{\alpha}}$ and the components $x_{1i},\dots,x_{\alpha i}$ of each vector $X_i$ are given in binary coding. The dimension~$\alpha$ may depend on the specific problem input.

Suppose that for a problem~$\Pi$ there exists a corresponding multi-objective problem~$P$ and an exact simplified DP algorithm defined by expression~(\ref{eqn:ext}). This algorithm works in~$n$ phases and for each $i=1,\dots,n$ the set of functions~${\mathcal F}_i$ and the function~$H_i$ do not depend on any input vectors other than~$X_i$. Besides that, $\Space \subset \Space' = {\mathbb N}_0^{\beta}$, where dimension~$\beta$ is fixed for~$\Pi$ and does not depend on a particular input~${\bf x}$. The assumption that elements of~$\mathcal{S}'$ are integer vectors will be essential in this section because each component of a state will actually be a quantitative parameter and will be subject to scaling. It is sometimes possible, however, to move from integer components to reals using the approach from~\cite{CERSW}.

The reduction from~$\Pi$ to~$P$, according to Section~\ref{sec:dynamic_programming}, implies that the Pareto-optimal set of~$P$ can be efficiently transformed into a solution to the problem~$\Pi$. Now let us suppose additionally that any $S\in {\mathcal S}_n$ can be mapped to some~$y(S) \in \mbox{\it Sol}_{\bf x}$ and there is a function~$G:{\mathbb N}_0^{\beta} \to {\mathbb N}_0$ such that $m_{\bf x}(y(S))=G(S)$.

The assumption that the simplified DP algorithm described in Section~\ref{sec:dynamic_programming} provides an exact solution to~$\Pi$ may be expressed formally:
\begin{equation} \label{eq:min}
\mbox{\it OPT}({\bf x})=\min\{G(S): S \in {\cal S}_n\},
\end{equation}
if $\Pi$ is a minimization problem, or alternatively
\begin{equation} \label{eq:max}
\mbox{\it OPT}({\bf x})=\max\{G(S): S \in {\cal S}_n\},
\end{equation}
if $\Pi$ is a maximization problem.

The function~$y(S)$ is usually computed by means of a standard backtracking procedure (see e.g.~\cite{CLRC01}, Chapter~15). A general description of such a procedure is beyond the scope of the paper since the details of reduction from problem~$\Pi$ to~$P$ are not considered here.

Suppose a {\em degree vector}~$D=(d_1,\dots,d_{\beta})\in {\mathbb N}_0^{\beta}$ is defined for~$\Pi$. Then, given a real value $\Delta >1$ we say that $S=(s_1,\dots,s_{\beta})$ is {\em $(D,\Delta)$-close} to $S'=(s'_1,\dots,s'_{\beta})$, if
$$
\Delta^{-d_{\ell}} s_{\ell} \le s'_{\ell} \le \Delta^{d_{\ell}} s_{\ell}, \quad
 {\ell}=1,\dots,\beta.
$$
Let us denote by~${\cal L}_0$ the set of indices $1 \le \ell \le \beta$ such that $d_{\ell}=0$, and let ${\cal L}_1=\{1,\dots,\beta\} \backslash {\cal L}_0$.

The main tool to exclude unpromising states in a DP-based FPTAS~\cite{Woeginger2000} is the quasi-linear order $\preceq_{\mathrm{qua}}$, which is an extension of a partial order $\preceq_{\mathrm{dom}}$, \ie if $S \preceq_{\mathrm{dom}} S'$ then $S \preceq_{\mathrm{qua}} S'$ for any $S,S'\in {\mathbb N}_0^{\beta}$. For the sake of compatibility with~\cite{Woeginger2000}, we will limit the consideration to the case where~$\preceq_{\mathrm{dom}}$ is a partial order, rather than a more general partial quasi-order as in Sections~\ref{sec:dynamic_programming} and~\ref{sec:evolutionary_algorithms}. This restriction is not significant w.\,r.\,t. applications of the framework, although most likely the results of~\cite{Woeginger2000}, as well as our results below, hold for the partial quasi-orders as well.

At each phase $i,\ i=1,\dots,n$, in DP$_{\Delta}$ only those states~$S$ may be excluded that are dominated in terms of~$\preceq_{\mathrm{qua}}$ by one of the other obtained states~$S'$, provided that $S'$ is $(D,\Delta)$-close to $S$.

Note that for any instance ${\bf x}$ the partial order $\preceq_{\mathrm{dom}}$ on the final sets ${\cal S}_1,\dots,{\cal S}_n$ may be represented by a finite number of criteria~$g_1,\dots,g_d$ of a corresponding instance of the problem~$P$ so that the Pareto-dominance relation is equivalent to $\preceq_{\mathrm{dom}}$ on this set.

A problem $\Pi$ is called {\em DP-benevolent} if besides C.1 and C.2, the following conditions C.1$'$,C.2$'$,C.3 and C.4 hold:\\

Condition C.1$'$. For any $\Delta>1$, $S,S' \in {\mathbb N}_0^{\beta}$ and $F \in {\cal F}_i, i=1,\dots,n$, if $S$ is $(D,\Delta)$-close to $S'$ and $S \preceq_{\mathrm{qua}} S'$, then either $F(S) \preceq_{\mathrm{qua}} F(S')$ and $F(S)$ is $(D,\Delta)$-close
to $F(S')$, or $F(S) \preceq_{\mathrm{dom}} F(S')$.\\

Condition C.2$'$. For any $\Delta>1$, $S,S' \in {\mathbb N}_0^{\beta}$ and $i=1,\dots,n$, if $S$ is $(D,\Delta)$-close to $S'$ and $S \preceq_{\mathrm{qua}} S'$, then $H_i(S') \le H_i(S)$.\\

Condition C.3. A value $\gamma \in {\mathbb N}_0$ exists, depending only on $G$ and $D$, such that for any $\Delta>1$ and $S,S' \in {\mathbb N}_0^{\beta}$,\\

(i) if $S$ is $(D,\Delta)$-close to $S'$ and $S \preceq_{\mathrm{qua}} S'$, then $G(S') \le \Delta^{\gamma} G(S)$ in the case of minimization, and $\Delta^{-\gamma} G(S) \le G(S')$ in
the case of maximization problem,\\

(ii) if $S \preceq_{\mathrm{dom}} S'$, then $G(S') \le G(S)$ in the case of minimization, and $G(S') \ge G(S)$ in the case of maximization problem.\\

Condition C.4.

(i) The functions $F \in {\cal F}_i$, $H_i, i=1,\dots,n$ and $G$, as well as the relation $\preceq_{\mathrm{qua}}$ are computable in time polynomially bounded in the input length.

(ii) $|{\cal F}_i|, \ i=1,\dots,n$ is polynomially bounded in input length.

(iii) ${\cal S}_0$ is computable in time polynomially bounded in input length.

(iv) A polynomial~$\pi_1(n,\log_2 |{\bf x}|)$ exists, such that all coordinates of any element $S \in {\cal S}_i$, $i=1,\dots,n$ are integer numbers bounded by $e^{\pi_1(n,\log_2 |{\bf x}|)}$. Besides that, for all $\ell\in {\cal L}_0$, the cardinality of the set of values that such a coordinate can take $|\{s_{\ell}: (s_1,\dots,s_{\ell},\dots,s_{\beta}) \in {\cal S}_i\}|$ is bounded by a  polynomial~$\pi_2(n,\log_2 |{\bf x}|)$.\\

\paragraph{Example: knapsack problem} We can verify the DP-benevolence conditions for the knapsack
problem as a simple illustrating example. Let the problem input, the DP states and the sets of mappings ${\mathcal F}_i, \ i=1,\dots,n$, as well as functions~$H_i$ be defined as in Section~\ref{sec:DP_examples}. Besides that, $G(S) \equiv s_2$ for all~$S=(s_1,s_2)\in {\mathcal S}_n$ and the degree vector is $D=(1,1)$.

A proper linear quasi-order~$\preceq_{\mathrm{qua}}$ that suits the partial order~$\preceq_{\mathrm{dom}}$ defined in Section~\ref{sec:DP_examples} for the knapsack problem is not known to us. Instead, we can consider the following relations~$\preceq_{\mathrm{qua}}$ and $\preceq_{\mathrm{dom}}$: let $S \preceq_{\mathrm{qua}} S'$ iff $s_1 \ge s'_1$, where $S=(s_1,s_2), \ S'=(s'_1,s'_2)$ and let $\preceq_{\mathrm{dom}}$ be the trivial partial order, \ie $S \preceq_{\mathrm{dom}} S'$ iff $S = S'$. (For an example of a DP-benevolent problem with non-trivial $\preceq_{\mathrm{dom}}$ see the problem of minimizing total late work on a single machine~\cite{Woeginger2000}.)

The statements in Conditions C.1, C.2, and C.3(ii) are fulfilled since $\preceq_{\mathrm{dom}}$ is trivial. The function $G(s_1,s_2)\equiv s_2$ satisfies Condition~C.3(i), which can be verified straightforwardly, assuming $\gamma=1$. To see that Condition C.4 holds, consider a polynomial $\pi_1(n,\log_2 |{\bf x}|) = \ln(2^{|{\bf x}|})$, which ensures that $ \max\{s_{\ell}\in {\mathcal S}_i  | \ell=1,2, \ i=1,\dots,n\} \le \max\{\sum_{i=1}^n p_i, W\} \le 2^{|{\bf x}|} = e^{\pi_1(n,\log_2 |{\bf x}|)}. $

Conditions C.1' and C.2' hold because the functions $F_i$, $\mbox{id}$ and $H_i$ at any phase~$i$ just sum the arguments with given non-negative constants. Indeed, consider e.g. the function $F_i(s_1,s_2)=(s_1+w_i, s_2+p_i)$. Here for any $\Delta>1$, if $s_{\ell}/\Delta \le s'_{\ell} \le \Delta s_{\ell}, \ \ell=1,2,$ then $(s_1+w_i)/\Delta \le s'_1+w_i \le \Delta (s_1+w_i)$ and $(s_2+p_i)/\Delta \le s'_2+p_i \le \Delta (s_2+p_i)$, therefore $F_i(s_1,s_2)$ is $(D,\Delta)$-close to $F_i(s'_1,s'_2)$. Besides that, adding a constant to~$s_1$ does not change the order $\preceq_{\mathrm{qua}}$. The functions $\mbox{id}$ and $H_i$ are treated analogously.

The other problems considered in Section~\ref{sec:DP_examples} either do not admit FPTAS unless P$=$NP (the TSP), or they are solvable in time which is polynomially bounded in the input length and thus do not require FPTAS (the SSSP and the APSP problems).

\subsection{Fully polynomial-time approximation scheme}
\label{subsec:fptas}

To identify subsets of states which are $(D,\Delta)$-close to each other, the algorithm DP$_{\Delta}$ employs a partition of the set of states into {\em $\Delta$-boxes} (defined below). This partition allows to discard ``close'' states analogously to discarding of $(1+\varepsilon)$-dominated solutions which is used in multi-objective optimization for approximation of Pareto-set (see e.g.~\cite{Horoba2009}). The main difference is that in our case the states are compared on the basis of their components, rather than the components of the vector of objectives. Note that usage of a quasi-linear order~$\preceq_{\mathrm{qua}}$ in DP$_{\Delta}$ will make $(D,\Delta)$-closeness only a necessary condition for discarding states from consideration.

Let~$L$ be a sufficiently large value, chosen for~${\bf x}$ and for any required precision~$\varepsilon \in (0,1)$ (a specific definition of~$L$ will be discussed later). To describe the algorithm DP$_{\Delta}$ let us consider a family of parallelepipeds that constitute a partition of the set $B(L,\Delta)={\mathbb N}_0^{\beta} \cap [0,\Delta^L]^{\beta}$:
$$
\{{\cal B}_{(k_1,\dots,k_{\beta})} : k_{\ell}=0,\dots,L, \ \ell=1,\dots,\beta\},
$$
where ${\cal B}_{(k_1,\dots,k_{\beta})}$ contains all integer points $S=(s_1,\dots,s_{\beta}) \in {\mathbb N}_0^{\beta}$, such that:
\begin{equation}
s_{\ell} \in
 \left\{
  \begin{array}{ll}
  0, & \mbox{ if } k_{\ell}=0,\\
  \mbox{[}\Delta^{k_{\ell}-1}, \Delta^{k_{\ell}}-1 \mbox{]}, &  \mbox{ if } 0<k_{\ell} <L,\\
  \mbox{[}\Delta^{k_{\ell}-1}, \Delta^{k_{\ell}}   \mbox{]}, &  \mbox{ if } k_{\ell} = L,
  \end{array}
 \right.
\end{equation}
for all $\ell \in {\cal L}_1$ and
$$
s_{\ell}=k_{\ell},
$$
for all $\ell \in {\cal L}_0$. Thus defined parallelepipeds are called $\Delta$-boxes below.\\

\begin{algorithm}[t]
    \caption{DP$_{\Delta}$ for~$\Pi$}
    \begin{algorithmic}[1]
        \STATE $\mathcal{T}_{0} \leftarrow \mathcal{S}_{0}$
        \FOR{$i=1$ to $n$}
            \STATE $\mathcal{T}_{i} \leftarrow \emptyset$
            \FOR{$S \in \mathcal{T}_{i-1}$ and $F \in \mathcal{F}_{i}$}
                \STATE let ${\cal B}_{(k_1,\dots,k_{\beta})}$ be
                the $\Delta$-box containing $F(S)$
                \IF{$H_i(F(S)) \leq 0$ and
                     $
                     \nexists S' \in \mathcal{T}_{i} \cap {\cal
B}_{(k_1,\dots,k_{\beta})} \colon F(S)
                        \preceq_{\mathrm{qua}} S'
                     $}
                    \STATE $\mathcal{T}_{i} \leftarrow (\mathcal{T}_{i} \setminus \{S' \in \mathcal{T}_{i} \cap {\cal
B}_{(k_1,\dots,k_{\beta})} \mid S' \prec_{\mathrm{qua}} F(S)\})
\cup \{F(S)\}$
                \ENDIF
            \ENDFOR
        \ENDFOR
        \STATE find $S^* \in {\cal T}_n$ such that
$$
    G(S^*)=
 \left\{
  \begin{array}{ll}
  \min \{G(S): S \in {\cal T}_n\} &\mbox{in case of minimization,}\\
  \max \{G(S): S \in {\cal T}_n\} &\mbox{in case of maximization}
  \end{array}
 \right.
$$
        \RETURN $y(S^*)$

    \end{algorithmic}\label{alg:FPTAS}
\end{algorithm}

Algorithm~\ref{alg:FPTAS} was suggested in~\cite{Woeginger2000} where it was proven to constitute an FPTAS with $\Delta$ and $L$ chosen as follows
\begin{equation}\label{choose_Delta}
\Delta=1+\frac{\varepsilon}{2\gamma n},
\end{equation}
\begin{equation}\label{choose_L}
L=\left\lceil \frac{\pi_1(n, \log_2 |{\bf x}|)}{\ln \Delta} \right \rceil.
\end{equation}
Equations~(\ref{choose_Delta}) and (\ref{choose_L}) ensure~$L$ is polynomially bounded in size of the input and in $1/\varepsilon$.

\subsection{Fully Polynomial-Time Randomized Approximation Scheme}

A family of randomized algorithms over all factors $0<\varepsilon<1$ with polynomially bounded running times in problem input size~$|{\bf x}|$ and in $1/\varepsilon$ that computes $(1+\varepsilon)$-approximate solutions with probability at least 3/4 is called a {\em fully polynomial-time randomized approximation scheme} (FPRAS)~\cite{JS93}. The constant 3/4 in the definition of FPRAS for optimization problems may be replaced by any other constant from the interval~(0,1).

The DP-based EA framework proposed in Section~\ref{sec:evolutionary_algorithms} may be modified to obtain an evolutionary FPRAS for DP-benevolent problems.

Now a new relation $\preceq_{\mathrm{\Delta}}$ is defined to substitute $\preceq_{\mathrm{dom}}$ in Algorithm~\ref{alg:ea}. Let us introduce the following relation: $(i,S) \preceq_{\mathrm{\Delta}} (i',S')$, iff $H_i(S) >0$ or the following three conditions hold:
\begin{enumerate}
\item[a)] $i=i'$\\
\item[b)] there exist such $k_1,\dots,k_{\beta}$ that $S,S' \in {\cal B}_{(k_1,\dots,k_{\beta})}$\\
\item[c)] $S \preceq_{\mathrm{qua}} S'$.
\end{enumerate}
The EA using this relation is denoted EA$_{\Delta}$ in what follows.


For an arbitrary $S \in {\cal S}_i$ let $\theta(i,S), \ i=1,\dots, n$, be the first iteration number, when an individual~$(i,T)$ was added into population, such that:
\begin{itemize}
\item[(i)] $T$ is $(D,\Delta^i)$-close to $S$ and
\item[(ii)] $S \preceq_{\mathrm{qua}} T$.
\end{itemize}
 In all iterations following~$\theta(i,S)$ the population will contain an individual~$T$ that satisfies the conditions (i) and (ii) as well.

The following lemma indicates that for any non-dominated $S \in {\cal S}_i$, in a number of iterations that is on average polynomially bounded in $|{\bf x}|$  and $1/\varepsilon$, an individual~$(i,T)$ will be obtained such that $T$ is $(D,\Delta^i)$-close to $S$ and $S \preceq_{\mathrm{qua}} T$. The proofs of the lemma and the theorem below are provided in~\cite{Eremeev10} but since this publication might be difficult to access, we reproduce the proofs here.

\begin{lemma} \label{EAlemma4_8}
Let~$\Pi$ be DP-benevolent with dimension $\beta$. Then for any stage $i=0,\dots,n$, any non-dominated state~$S$ in ${\cal S}_i$ and $L$ chosen as defined in Equation \ref{choose_L} it holds that
$$
E[\theta(i,S)] \le n(L\pi_2(n, \log_2 |{\bf x}|))^{\beta} \cdot \sum_{k=1}^i |\mathcal{F}_k|.
$$
\end{lemma}

{\bf Proof.} Let us use induction on~$i$. For $i=0$ the statement holds trivially. Consider any state~$S$ which is non-dominated in ${\cal S}_{i}$. Suppose $i>0$ and the statement holds for~$i-1$.

Lemma~4.7 in~\cite{Woeginger2000} implies that there exists a state~$S^{\#}$ non-dominated in ${\cal S}_{i-1}$ and a mapping~$F^{\#} \in {\cal F}_i$, such that $F^{\#}(S^{\#}) =S$. Note that the induction hypothesis gives an upper bound on expectation of~$\theta(i-1,S^{\#})$, which is the expected number of iterations until an individual $(i-1,T^{\#})$ is obtained, such that $T^{\#}$ is $(D,\Delta^{i-1})$-close to $S^{\#}$ and $S^{\#} \preceq_{\mathrm{qua}} T^{\#}$. Again, let the mutation that applies $F^{\#}$ to an individual $(i-1,T^{\#})$ be called a successful mutation.

In view of C.2$'$ condition,
$$
H_{i}(F^{\#}(T^{\#})) \le H_{i}(S) \le 0,
$$
and by C.1 and C.1$'$, either (a) $F^{\#}(T^{\#})$ is $(D,\Delta^{i-1})$-close to $S$ and $S \preceq_{\mathrm{qua}} F^{\#}(T^{\#})$, or (b) $S \preceq_{\mathrm{dom}} F^{\#}(T^{\#})$.

In case (a), after a successful mutation, the population will contain the element $(i,F^{\#}(T^{\#}))$, or some other element $(i,T')$ such that $T'$ belongs to the $\Delta$-box~${\cal B}_{(k_1,\dots,k_{\beta})}$, which also contains $F^{\#}(T^{\#})$ and besides this $F^{\#}(T^{\#}) \preceq_{\mathrm{qua}} T'$. After this mutation the population will contain an individual $(i,T)$, such that $T$ is $(D,\Delta)$-close to $F^{\#}(T^{\#})$ and $F^{\#}(T^{\#}) \preceq_{\mathrm{qua}} T$. Now since $F^{\#}(T^{\#})$ is $(D,\Delta^{i-1})$-close to~$S$, by the definition of closeness, $T$ is $(D,\Delta^s)$-close to $S$. Besides that, $S \preceq_{\mathrm{qua}} F^{\#}(T^{\#}) \preceq_{\mathrm{qua}} T$, consequently, $S \preceq_{\mathrm{qua}} T$. Thus, in case (a), successful mutation ensures presence of the required representative for~$S$ in population on stage~$i$.

In case (b), a successful mutation will yield the individual
$(i,S)$, since $S$ is a non-dominated state, and $S
\preceq_{\mathrm{dom}} F^{\#}(T^{\#})$.
After such a mutation, the population will contain an individual $(i,T)$, such that $T$ is $(D,\Delta)$-close to $S$ and $S \preceq_{\mathrm{qua}} T$. Obviously, $T$ is also $(D,\Delta^i)$-close to $S$ then.

To complete the proof it remains to estimate the expected number of mutation attempts~$\theta^*$ until a successful mutation occurs, conditioned that an individual $a^{\#}={(i-1,T^{\#})}$ belongs to the current population~${\cal P}$. Note that the probability of a successful mutation is
$$
p^*=\left(n\cdot |\{(i-1,S')\in {\cal P}\}| \cdot |\mathcal{F}_i| \right)^{-1},
$$
at the same time,
\begin{equation}\label{upperbound}
|{\cal P}| = \sum_{i'=1}^n |\{(i',S')\in {\cal P}\}| \le \sum_{i'=1}^n |\{(k_1,\dots,k_{\beta}): {\cal B}_{(k_1,\dots,k_{\beta})}\cap {\cal S}_{i'}\ne \emptyset\}|.
\end{equation}

Consider a single term in the right-hand side of inequality~(\ref{upperbound}) with any fixed~$i'$. For each~$\ell\in {\cal L}_1$ the index~$k_{\ell}$ may take at most~$L$ different values. Besides that, in view of condition~C4~(iv), for each $\ell'\in {\cal L}_0$ the coordinate~$s_{\ell'}$ characterizing the states from the set~${\cal S}_{i'}$ may take at most $\pi_2(n, \log_2 |{\bf x}|)$ values.

Thus, the right-hand side of inequality~(\ref{upperbound}) can not exceed
$$
nL^{|{\cal L}_1|} \pi_2(n, \log_2 |{\bf x}|)^{|{\cal L}_0|} \le n(L\pi_2(n, \log_2 |{\bf x}|))^{\beta}.
$$
The statement of the lemma for phase~$i$ follows from the fact that $E[\theta(i,S)]=E[\theta(i-1,T^{\#})]+1/p^*$. \qed\\

The bound on $E[\theta(i,S)]$ obtained in Lemma~\ref{EAlemma4_8} is used to choose the stopping criterion for the algorithm EA$_{\Delta}$. Let the algorithm terminate after
\begin{equation}\label{stop_crit}
\tau=4n(L\pi_2(n, \log_2 |{\bf x}|))^{\beta} \sum_{i=1}^n
|\mathcal{F}_i|
\end{equation}
iterations.

\begin{theorem} \label{EAtheorem}
If the problem~$\Pi$ is DP-benevolent, then the family of algorithms EA$_{\Delta}$ where $\Delta$ and $L$ are chosen according to~(\ref{choose_Delta}) and (\ref{choose_L}), using the stopping criterion~(\ref{stop_crit}) gives an FPRAS.
\end{theorem}

{\bf Proof.} In view of (\ref{eq:min}) and C3~(ii), there exists a non-dominated state $S^* \in {\cal S}_n$, such that $OPT({\bf x})=G(S^*)$. By Lemma~\ref{EAlemma4_8}, on average within at most $n(L\pi_2(n, \log_2 |{\bf x}|))^{\beta} \cdot \sum_{i=1}^n |\mathcal{F}_i|$ iterations of EA$_{\Delta}$, a population will be computed, containing an individual~$(n,T^*)$, such that $T^*$ is $(D,\Delta^n)$-close to $S^*$ and $S^* \preceq_{qua} T^*$.

Let us first consider the case where~$\Pi$ is a minimization problem. By condition C3~(i):
$$
G(T^*) \le \Delta^{\gamma n} G(S^*) = \left(1+\frac{\varepsilon}{2\gamma n}\right)^{\gamma n} OPT({\bf x}) \le (1+\varepsilon) OPT({\bf x}).
$$
The latter inequality follows from the observations that $\gamma n \ge 1$, $(1+\frac{\varepsilon}{2\gamma n})^{\gamma n}$ is a convex function in~$\varepsilon$ on the interval $\varepsilon \in [0,2]$, and the indicated inequality holds for both endpoints of this interval. In the case of maximization problem~$\Pi$ analogously we obtain $ G(T^*) \ge (1+\varepsilon)^{-1} OPT({\bf x}).$

Finally, by means of backtracking, a $(1+\varepsilon)$-approximate solution~$y(T^*)$ may be computed efficiently.

Execution of EA$_{\Delta}$ with stopping criterion~(\ref{stop_crit}), according to the Markov inequality, does {\em not} yield a $(1+\varepsilon)$-approximate solution with probability at most $1/4$.

Finally, by condition C.4, the runtime of each iteration of the EA$_{\Delta}$ is polynomially bounded in the input length and in $1/\varepsilon$. Summing up the observed facts, we conclude that the proposed family of the algorithms constitutes an FPRAS. \qed\\

\section{Conclusions}
\label{sec:conclusions}

We have examined how to choose a representation for an evolutionary algorithm such that it obtains the ability to carry out dynamic programming. Based on a general framework for dynamic programming we have given a framework for evolutionary algorithms that have a dynamic programming ability and analyzed the optimization time of such an algorithm depending on the corresponding dynamic programming approach. By considering well-known combinatorial optimization problems, we have shown that our framework captures most of the known DP-based evolutionary algorithms and allows to treat other problems.

\subsection*{Acknowledgements}

The authors would like to thank the organizers of the Theory of Evolutionary Algorithms seminars at Schloss Dagstuhl, where this research was started. Also, the authors are grateful to Alexander Spirov for the helpful comments on mutation mechanisms. The research was supported in part by Presidium RAS (fundamental research program~2, project~227).



\appendix

\section{Bellman Principle for Single-Objective Problems} \label{app:bellman}

In this appendix, we describe the Bellman optimality principle in terms of the DP method defined by recurrence~(\ref{eqn:ext}). Consider a single-objective maximization problem~$\Pi$. Let $2\le{\beta} \in \Nats$ be a constant and ${\cal S} \subseteq \Reals^{\beta}$, so that the first component~$s_1$ of a state~$S \in {\cal S}$ characterizes a quality of the state in some sense.

The Bellman principle applies to a DP algorithm for~$\Pi$ if the following statement holds. Suppose that starting from some state~$S_{0}^* \in {\cal S}_0$, a sequence of ``decisions'' $F_1\in {\cal F}_1,\dots, F_n \in {\cal F}_n$ leads to an optimal solution for~$\Pi$. Let us denote $S_i^*=F_i(F_{i-1}(...F_{1}(S_{0}^*)...)) \in {\cal S}_i,\ i=1,\dots,n$. Then for any particular state~$S_i^*=(s_{1i}^*,\dots,s_{\beta i}^*)$, the subsequence $F_1,\dots,F_i$ is an optimal policy for reaching the set of states coinciding with $S_i^*$ in components $s_2,\dots,s_{\beta}$. By an optimal policy here we mean that for any sequence $F'_1\in {\cal F}_1,\dots, F'_i \in {\cal F}_i$ starting with some~$S'_{0} \in {\cal S}_0$, such that $S'_k=F'_k(F'_{k-1}(...F'_{1}(S'_{0})...)) \in {\cal S}_k,\ k=1,\dots,i,$ and $S'_i=(s'_{1i},s_{2i}^*,\dots,s_{\beta i}^*),$ holds $s'_{1i} \le s_{1i}^*$.

If the Bellman principle applies to a DP algorithm, then for any $s_{2},\dots,s_{\beta}$ it is possible to keep only one state which dominates all states in the subset $\{S' \in {\cal S}_i \ : \ s'_{2}=s_{2},\dots, s'_{\beta}=s_{\beta} \}$ without a risk to loose optimality of the DP algorithm result.

\section{Genetic Mechanisms Corresponding to the
Mutation Proposed in the EA} \label{app:mutation}

The mutation operator proposed in the EA in Section~\ref{subsec:modules} is a special case of the point mutation, where a gene~$A_i$ subject to change is selected as the first gene which has never been mutated so far (\ie the first ``undefined'' gene). Such type of mutation may be imagined in a biological system as follows.

Suppose that for each phase~$i, \ i=1,\dots,n$, there is a ``controlling'' gene~$B_i$. The required localization of mutations in gene~$A_i$, when~$A_i$ is the first ''undefined'' gene, is caused by insertion of some mobile DNA sequence~$C_i$ (e.g. a transposon, see~\cite{Sherratt}), that can enter the locus of gene~$A_i$, and only this locus. We can additionally assume that a mobile element~$C_i$ is produced if and only if the gene~$B_i$ is active (\ie $B_i$ is subject to transcription in the parent individual). Besides that, we can assume that gene~$A_i$ in the ``undefined'' condition is silencing the transcription of gene~$B_{i+1}$, but any mutated state of gene~$A_i$ activates the transcription of gene~$B_{i+1}$ and silences the gene~$B_{i}$.

Then one can assume that in the $i$-th generation, $i=1,\dots,n$, only the gene~$B_{i}$ is active among $B_{1},\dots,B_{n}$, provided that initially only the gene~$B_{1}$ was active. At the same time, in the $i$-th generation, $i=1,\dots,n$, the insertion mutations occur only in the gene~$A_{i}$.

In nature, an example of a mutually exclusive genes activation
is observed in malaria parasite Plasmodium falciparum. The
transitions from one variant of a gene to another one depend on
the currently active gene variant~\cite{Horr04}.



\bibliographystyle{abbrv}
\bibliography{bibtex}







\end{document}